\documentclass[runningheads]{llncs}
\usepackage{graphicx}
\usepackage{multirow}
\graphicspath{ {figures/} }
\usepackage{todonotes}
\usepackage{wrapfig}
\usepackage{subcaption}
\usepackage{listings}
\usepackage{xcolor}
\lstdefinestyle{SPARQLdeco}{language=Python,  
                            columns=fullflexible,  
                            xleftmargin=6mm,                    
                            linewidth=4.8in,
                            breaklines=true,
                            escapeinside={(*}{*)},
                            keepspaces,                         
                            backgroundcolor=\color{black!4!white}, 
                            numbers=left,                       
                            numberstyle=\scriptsize,            
                            frame=l,                            
                            basicstyle=\scriptsize\ttfamily,  
                            keywordstyle=\bfseries}             

\begin{document}
%
\title{Drugs4Covid: Drug-driven Knowledge Exploitation based on Scientific Publications}

\titlerunning{Drugs4Covid}
%
\author{Carlos Badenes-Olmedo\inst{1} \and
David Chaves-Fraga\inst{1} \and
Mar\'ia Poveda-Villal\'on\inst{1} \and
Ana Iglesias-Molina\inst{1} \and
Pablo Calleja\inst{1} \and
Socorro Bernardos\inst{1} \and
Patricia Mart\'in-Chozas\inst{1} \and
Alba Fern\'andez-Izquierdo\inst{1} \and
Elvira Amador-Dom\'inguez\inst{1} \and
Paola Espinoza-Arias\inst{1} \and
Luis Pozo\inst{1} \and
Edna Ruckhaus\inst{1} \and
Esteban Gonz\'alez-Guardia\inst{1} \and
Raquel Cedazo\inst{2} \and
Beatriz L\'opez-Centeno\inst{3} \and
Oscar Corcho\inst{1}
}
\authorrunning{Badenes-Olmedo et al.}
%
\institute{Ontology Engineering Group, Universidad Polit\'ecnica de Madrid, Boadilla del Monte, Spain \and
High School of Technical Industrial and Design Engineering, Universidad Polit\'ecnica de Madrid, Madrid, Spain \and
Subdirecci\'on General de Farmacia y Productos Sanitarios, Servicio Madrileño de Salud, Madrid, Spain 
}
\maketitle              
\begin{abstract}

In the absence of sufficient medication for COVID patients due to the increased demand, disused drugs have been employed or the doses of those available were modified by hospital pharmacists. Some evidences for the use of alternative drugs can be found in the existing scientific literature that could assist in such decisions. However, exploiting large corpus of documents in an efficient manner is not easy, since drugs may not appear explicitly related in the texts and could be mentioned under different brand names. Drugs4Covid combines word embedding techniques and semantic web technologies to enable a drug-oriented exploration of large medical literature. Drugs and diseases are identified according to the ATC classification and MeSH categories respectively. More than 60K articles and 2M paragraphs have been processed from the CORD-19 corpus with information of COVID-19, SARS, and other related coronaviruses. An open catalogue of drugs has been created and results are publicly available through a drug browser, a keyword-guided text explorer, and a knowledge graph.

\keywords{Ontology-based technologies \and NLP \and Bio-annotations \and Drugs-catalogue \and Knowledge Graph \and COVID-19.}
\end{abstract}

\section{Introduction}\label{sec:intro}

News on the scarcity of medicines to treat COVID-19 have appeared in many countries around the world over the past months. Due to the increasing demand during the peaks of infections worldwide and other problems associated to logistics, doctors and pharmacists were struggling to treat their patients. We confirmed this situation after a series of interviews with doctors, pharmacists and other people from regional Health Services. In the absence of sufficient medication for such a large amount of patients, practitioners started applying disused drugs that were available in hospital pharmacies and modifying the usual doses of those available. 

\begin{figure}[ht!]
\includegraphics[scale=0.45]{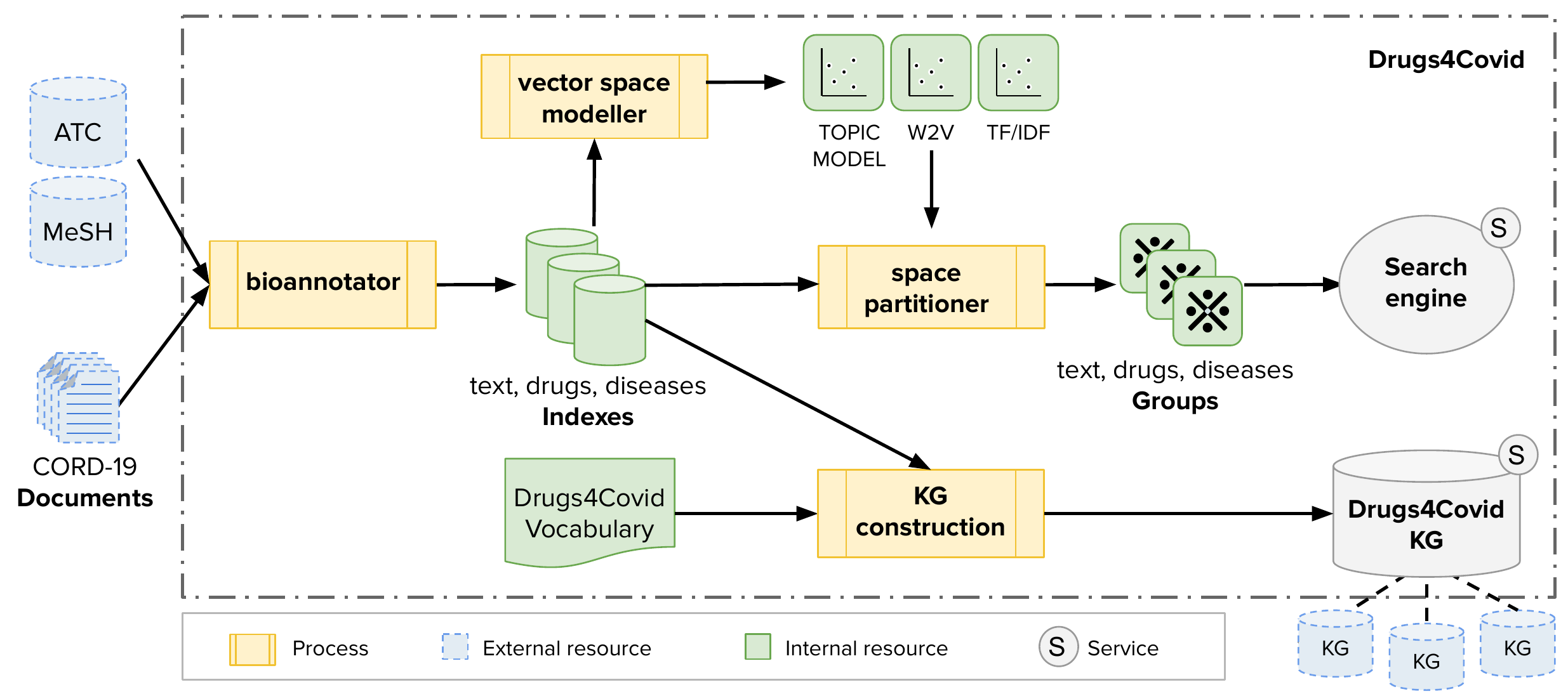}
\centering
\caption{Creation workflow of a search engine and a knowledge graph through annotations created from the CORD-19 dataset. }
\label{fig:complete-workflow}
\end{figure}

However, identifying which drugs can be used as replacements for others to treat this little-known disease and its associated symptoms is a difficult challenge. New experiments and results are continually being published, and people in charge of clinical protocols cannot keep up to date with all of them \cite{brainard2020}. This situation calls for solutions that help health care providers and researchers easily extract such knowledge from the enormous scientific corpus that is being created.

Several initiatives have emerged to bring together scientific publications in this domain. For instance, the COVID-19 Data Portal maintained by the EU, or those published by Humandata focused on COVID-19 cases around the world. Similarly, other research groups and universities have also published valuable datasets, such as the Open COVID-19 Data Working Group\footnote{\url{https://github.com/beoutbreakprepared/nCoV2019}}, the DisGeNET COVID-19 dataset\footnote{\url{https://www.disgenet.org/covid/diseases/summary/}} or the data gathered by John Hopkins University\footnote{\url{https://github.com/CSSEGISandData/COVID-19}}, among others.

The Allen Institute for Artificial Intelligence created the COVID-19 Open Research Dataset (CORD-19)~\cite{wang2020cord}. It is a continuously growing corpus with all publicly available COVID-19 and coronavirus-related research (e.g. SARS, MERS, etc.) in the last fifty years. This dataset can be used as a source of information to extract knowledge related to the disease. At the time of this study, it is composed of 23,428 open access articles from PubMed Central, 35,240 research articles from a corpus maintained by the World Health Organization (WHO), and 1,945 bioRxiv and medRxiv pre-prints.

Given that volume, a combined used of natural language processing with word embeddings techniques and knowledge extraction technologies can provide doctors and medical researchers with tools that make their work easier by structuring the information contained within the papers. Our goal is to automatically extract, organize and publish the drug-oriented knowledge from the medical literature needed to answer common questions posed by domain experts such as \textit{What are the effects of using chloroquine and hydroxychloroquine to treat COVID-19 patients? Have drugs that combine immunosuppressive and antimalarial activity with macrolide antibiotics been used? In which experiments have mefloquine and azithromycin been used and related to which diseases?} 


We processed more than 60k medical publications to discover relations amongst drugs and diseases. The tools created from our workflow help to quickly create domain-specific search engines and knowledge graphs (KG) (Fig. \ref{fig:complete-workflow}) over any corpus of scientific documents. Such techniques can be reused for other similar crisis in the future, and in any other situation where these tools could be valuable. All the work presented in this paper is available in a GitHub repository\footnote{\url{https://github.com/oeg-upm/drugs4covid19}}

\section{Related Work} \label{related work}

Due to the fast evolution of the SARS-COV-2, worldwide scientists have been working as fast as possible to provide tools that help health professionals treat patients during this pandemic. Most of the ongoing work has been shared inside research communities (e.g. W3C Semantic Web and W3C Healthcare and Life Sciences), in hackathons (e.g. the European Commission hosted \textit{EUvsVirus}\footnote{\url{https://www.euvsvirus.org}}, \textit{VenceAlVirus}\footnote{\url{https://vencealvirus.org}} in the region of Madrid), and in preliminary reports in blogs and archival services. Multiple datasets\footnote{\url{https://data.world/resources/coronavirus/}} have been provided for finding new insights about the novel coronavirus.

Semantic Web technologies have been traditionally used to publish many data sources in the biomedical domain such as \textit{Bio2RDF} that provides a network of linked data for the biomedical domain, \textit{DisGeNET RDF} that relates diseases and genes, or \textit{SNOMED Clinical Terms} that provides a clinical healthcare terminology. Some of this earlier work has been used by language technologies to add semantic annotations to scientific literature. The \textit{CORD19-Named Entities Knowledge} identified and disambiguated named entities in  CORD-19 using \textit{DBpedia Spotlight} and the \textit{NCBO BioPortal} annotator. Similarly, a Linked Data version of the CORD-19 data is provided by the CORD-19-on-FHIR project introducing semantic annotations on conditions, medications, and procedures. 

Some knowledge graphs have also been created such as the \textit{Knowledge4COVID-19}\footnote{\url{https://devpost.com/software/covid-19-kg}} that integrates knowledge from scientific literature and databases, and uses predictive models to derive interactions among drugs used in the treatment of COVID-19. Others are mainly focused on a collective generation of knowledge such as \textit{KG-COVID-19 Hub}\footnote{\url{https://github.com/Knowledge-Graph-Hub/kg-covid-19}} or the \textit{Covid Graph project}\footnote{\url{https://live.yworks.com/covidgraph}}. Both seek to build collaboratively a KG hub for COVID-19 so that developers can ingest new sources into this hub. And there are also initiatives focused on particular domains such as the \textit{SIB COVID-19 Integrated Knowledgebase} that is specialized on genetic information, and the \textit{COVID-19 by STKO Lab} that combines recorded cases and suspensions of commercial airline routes.  

However, none of these approaches provide sufficient information on how the knowledge is generated. When two drugs are related through an interaction in the knowledge graph, the information used to induce that kind of relation is missing and would be very relevant for its potential users. We believe that conclusions are as important as the data that led to them during the learning process. Hence, in this paper we describe a process to create a knowledge graph that, together with the diseases and drugs identified in the CORD-19 corpus, links to the texts (i.e paragraphs, sentences or even full articles) that support the inferred knowledge.

\section{Drug and Disease Annotations}
\label{bioannotations}
Searching for similar drugs and exploring major diseases covered by different publications are key activities when browsing medical papers. This manual know-\\ledge-intensive task becomes less tedious and even leads to unexpected relevant findings when applying unsupervised algorithms to help researchers. 

In this work we have addressed the problem of generating drug- and disease-based annotations for the documents within a large collection of research articles. We have created vectorial representations of each of them using state-of-the-art techniques based on word embeddings and probabilistic topic models. Then, diseases and drugs have been related to each other and suggested drugs that may be considered as replacements for others.  

More than 60k scientific papers were analyzed from the CORD-19 dataset. Around 5M sentences and 2M paragraphs were annotated with the drugs and diseases mentioned in them. A total of 6,400 diseases and 2,400 drugs were characterized to enrich the searches on the corpus and the knowledge graph. 

\subsection{Taxonomies used for annotation} \label{section:pharmacologic}
Textual searches on a collection like CORD-19 bring accuracy issues since the terminology associated with drugs and diseases varies between countries. A search based on an active substance should lead to the drugs that contain it. Such drugs can be traded under different brand names among countries, hence appearing with different names in different papers. For example, the active substance \textit{Enalapril/Hydrochlorothiazide} is distributed in Spain under the trade names \textit{Co-Renitec}, \textit{Crinoretic} or \textit{Dabond Plus}, while its trade name is \textit{Renidur} in Portugal, and \textit{Corodil} in Denmark. Thus, we need to unify drugs and diseases with codes that abstract them from their particular names or textual representations. 

We have used the trade names published by the Spanish Agency of Medicines and Health Products (AEMPS) and considered the Anatomical Therapeutic Chemical (ATC) classification system to annotate drugs. The ATC is supported by the World Health Organization (WHO) and widely used in hospital pharmacies to identify drug components. It groups active substances according to the organ or system on which they act and their therapeutic, pharmacological, and chemical properties. Drugs are classified into groups at five different levels. The first one corresponds to main groups, the second one to pharmacological or therapeutic subgroups, the third and the fourth one are chemical-pharmacological-therapeutic subgroups and the last one is the chemical substance.

Diseases have been annotated with the Medical Subject Headings (MeSH) vocabulary, which is used for indexing, cataloging, and searching biomedical information by the MEDLINE/PubMed article database, the NLM\'{}s books catalog, and ClinicalTrials.gov among others. This vocabulary is organized in a hierarchical, numbered tree structure that enables browsing from broader to narrower topics.

Our goal is not to annotate according to a single classification system, but to integrate different classification codes suggested by our users to allow cross-searching from their terminologies.

\subsection{Annotation techniques}
\label{sec:annotations}

\begin{table}[t]
\small
\centering
 \begin{tabular}{||c c c c||} 
 \hline
 \textbf{Top-Word} & \textbf{P01BA01}  & \textbf{P01BA02} & \textbf{J05AR10} \\ [0.5ex]
 \hline\hline
 1 & chloroquine & chloroquine & mer \\ 
 \hline
 2 & chikv & hydroxychloroquine & camel \\
 \hline
 3 & autophagy & glycosylation & cov \\
 \hline
 4 & endosome & analog & mild \\
 \hline
 5 & cathepsin & rheumatoid & lopinavir \\ [1ex] 
 \hline
\end{tabular}
\caption{Top5 words related to \textit{Chloroquine} (P01BA01), \textit{Hydroxychloroquine} (P01BA02) and \textit{Lopinavir/Ritonavir} (J05AR10) according to the PTM created from the active substances discovered in the CORD-19 corpus. }
\label{table:topic-models}
\end{table}

Common entities identified in medical literature are diseases, drugs and chemical names among others~\cite{tsai2006various,goulart2011systematic,song2018comparison,cho2019biomedical}. These named entities can also be used to discover relations between them. 
We created a pipeline to prepare the textual content from the CORD-19 dataset~\cite{wang2020cord} and index it together with some additional information in a document-oriented database. It is based on the librAIry framework~\cite{Badenes-Olmedo2017}, which provides a high-performance infrastructure for text mining tasks. We focused on the following data for this task: 
\begin{itemize}
  \item id: unique identifier for each research article.
  \item name: publication title.
  \item abstract: brief summary made by the author.
  \item text: full textual content of the article.
  \item url: online resource available from Allen AI repository.
\end{itemize}

The Named Entity Recognition (NER) task was performed combining different state-of-the-art methods such as scispaCy~\cite{neumann-etal-2019-scispacy} and CliNER~ \cite{boag2018cliner}. After they were tested to identify disorders, substances, trials and treatments from texts, we realized the methods did not unify the identification of drugs and diseases as described in the ATC or MeSH classifications. Sometimes they are identified as different drugs or diseases, since they are referred by different names, when in fact they correspond to the same element. To adjust this behavior we have developed Bio-NLP\footnote{\url{https://github.com/librairy/bio-nlp}}, a NER service built on top of scispaCy that uses the MeSH vocabulary and the ATC codes as gazetteers of terms respectively to identify named entities. 

\section{Drug and Disease Representational Models}\label{sec:models}
In order to establish relations among drugs or diseases, so as to improve searches on the corpus, we have created representational models capable of abstracting the entities from their textual forms and relating them to each other. 

\subsection{Identifying Related Texts}

Some text mining algorithms represent documents in a common feature space that abstracts away from the specific sequence of words used in each of them. That is, they treat documents as bags-of-words. With appropriate representations, they ease the analysis of relationships between documents, even when written using different vocabularies.


\begin{figure}[t]
\includegraphics[scale=0.23]{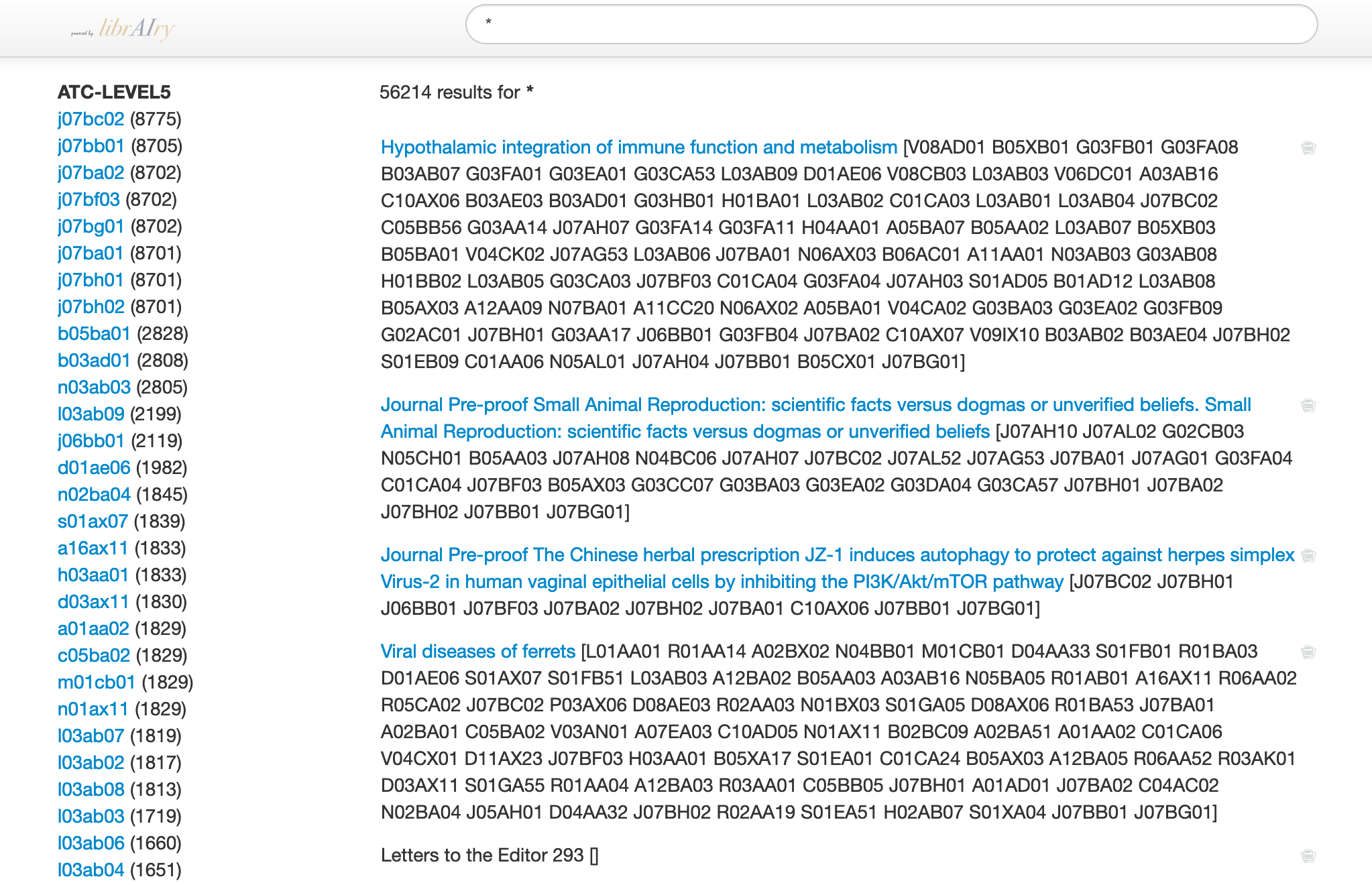}
\centering
\caption{Documents described by a PTM from the active substances (ATC-code level5) in the CORD-19 corpus and organized by similar topic distributions.}
\label{fig:covid19-explorer}
\end{figure}

\begin{itemize}
\item \textbf{Representational Model}: We have used topic models, which are widely used to uncover the latent semantic structure from text corpora. In particular, Probabilistic Topic Models (PTM) represent documents as a mixture of topics, where topics are probability distributions over words. This eases the exploitation of large document collections, since documents are mapped into a low-dimensional latent space (i.e they are described by a small number of topics). It also abstracts the representation of the documents from the words themselves. Topics emerge as density distributions on the vocabulary to define the dimensions of the vectors that represent the documents. We used librAIry\cite{Badenes-Olmedo2017} to create and fine-tune a PTM that describes the annotated drugs as probability distributions on the vocabulary of the corpus. Each topic matches an active substance (see Table \ref{table:topic-models}), avoiding distinctions between different drug names by country. The model is created and distributed as a Rest-API to facilitate its reuse.
\item \textbf{Similarity Metric}: Due to low storage cost and fast retrieval speed, hashing is a popular solution for the calculation of approximate nearest neighbors in the probability simplex space created from topic models. We opted for a density-based hierarchical hashing method \cite{Badenes-Olmedo2020} to compare the texts contained in CORD-19. This approach has proven to obtain high-precision results and can accommodate additional query restrictions.
\item \textbf{Space Partitioning}: Since documents are compared from sets of topics distributed in levels according to their relevance, the representational space can be organized in k-d trees. The most discriminating topics and levels define the branches of the tree and each text is organized among them according to its topic distributions. We created a web-based document browser\footnote{\url{https://librairy.github.io/covid19/explorer}} to navigate through the corpus using the active substances and content-based relations (Figure \ref{fig:covid19-explorer}).
\end{itemize}

\subsection{Finding Related Drugs}
Drugs were described by the diseases with which they are mentioned in texts (e.g sentence, paragraph, or full-text). Paragraphs showed an appropriate balance between generality and particularity to describe drugs, since their content is more strongly related than the full-text, and less incomplete than the sentences.



\begin{figure}[t]
\centering
\centering
	\begin{subfigure}{.4\linewidth}
	   \includegraphics[width=0.9\columnwidth]{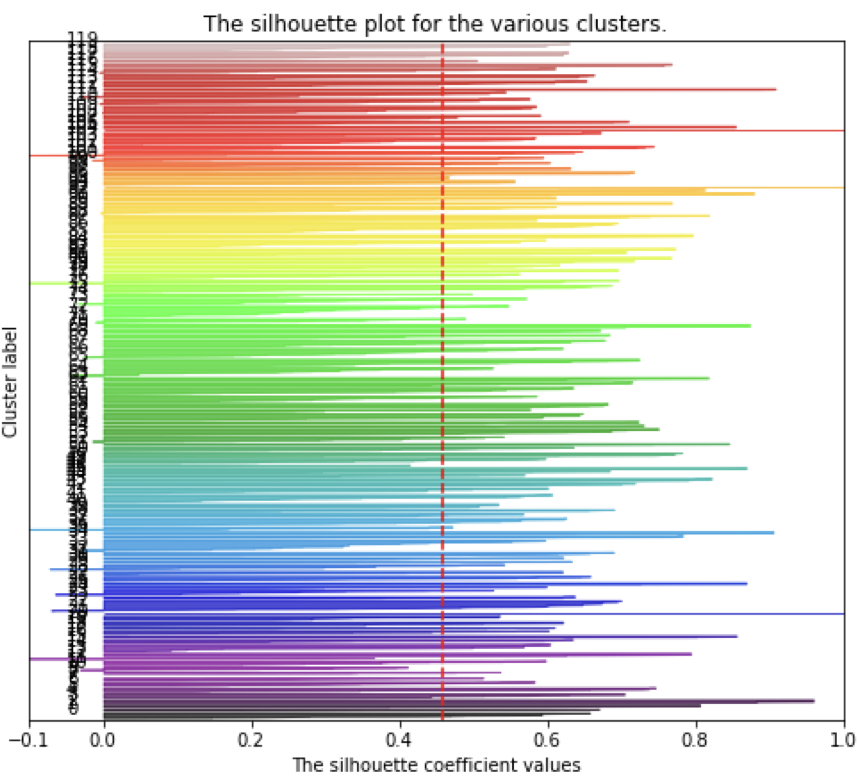}
	   \caption{Silhouette analysis}
       \label{fig:silouette120}
	\end{subfigure}
	\begin{subfigure}{.4\linewidth}
        \includegraphics[width=0.9\columnwidth]{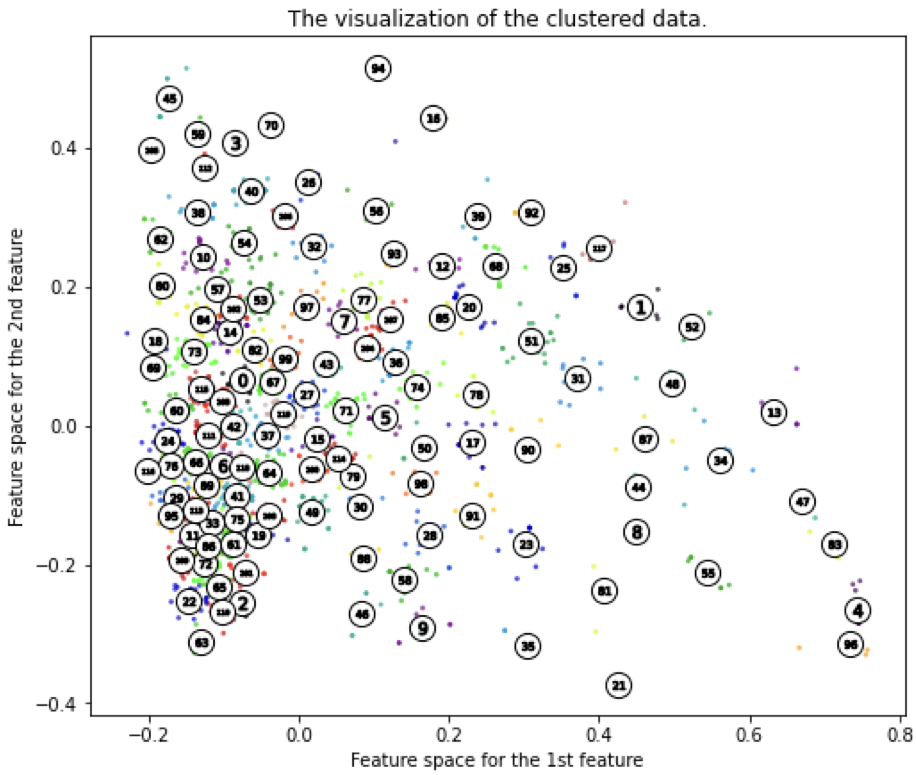}
        	\caption{Two-dimensional space}
            \label{fig:drugsCluster}
    	\end{subfigure}
\caption{Validation of the 120 clusters of drugs.}
\label{fig:silhouette-validation}
\end{figure}

\begin{itemize}
\item \textbf{Representational Model}: A drugs-diseases matrix was created with the times a disease appears in the same paragraph than a drug. Then, we used Term Frequency-Inverse Document Frequency (TFIDF) to transform drugs into numeric vectors. Two drugs described by TFIDF will be similar if they share rare, but informative, diseases.
\item \textbf{Similarity Metric}: Cosine similarity was adopted as the measure of interest as it is well-known and frequently-used in vector space models \cite{Shahmirzadi2018TextSI}.
\item \textbf{Space Partitioning}: Similar drugs are organized into the same groups to suggest their use as replacements. Since the number of groups is unknown, we created a dendrogram based on simple linkage from their vectorial representations. As a result we obtained a hierarchical aggregation of drugs that suggests an optimal number of groups according to the Silhouette coefficient. Several settings were tested and finally 120 clusters offered the best performance (Fig. \ref{fig:silhouette-validation}). We developed a Rest-API, Bio-API\footnote{\url{https://librairy.linkeddata.es/bio-api/replacements?keywords=chloroquine}}, that makes use of the Annoy\footnote{\url{https://github.com/spotify/annoy}} method to index drugs in a kd-tree. It allows queries like \textit{bio-api/replacements?keywords=chloroquine} to retrieve related drugs.
\end{itemize}

\subsection{Finding Related Diseases}
In order to measure the similarity amongst diseases, a three-step procedure was followed: 1) Creation of disease-focused word embedding models, 2) Generation of a terminology representing the whole corpus, used to align the compared diseases, 3) Disease similarity calculation from the identified key terms.

\begin{table*}
\centering
\resizebox{\textwidth}{!}{%
\begin{tabular}{||c|c|c|c|c||}
\hline
\multicolumn{5}{||c||}{\textbf{CORD-19 top 25 terms}}                                                                  \\ \hline
T cell          & respiratory tract  & viral replication & infected cell        & health care                \\ \hline
public health   & infectious disease & viral RNA         & cell lines           & RNA viruses                \\ \hline
amino acid      & immune responses    & epithelial cell   & respiratory syndrome & nucleic acid               \\ \hline
immune response & accute respiratory & E. coli           & virus infection      & acute respiratory syndrome \\ \hline
influenza virus & gene expression    & viral infection   & immune system        & respiratory viruses        \\ \hline
\end{tabular}%
}
\caption{Main terms discovered in the CORD-19 corpus}
\label{tab:terms}
\end{table*}

\begin{figure}
\centering
	\begin{subfigure}{.4\linewidth}
        \includegraphics[width=0.9\columnwidth]{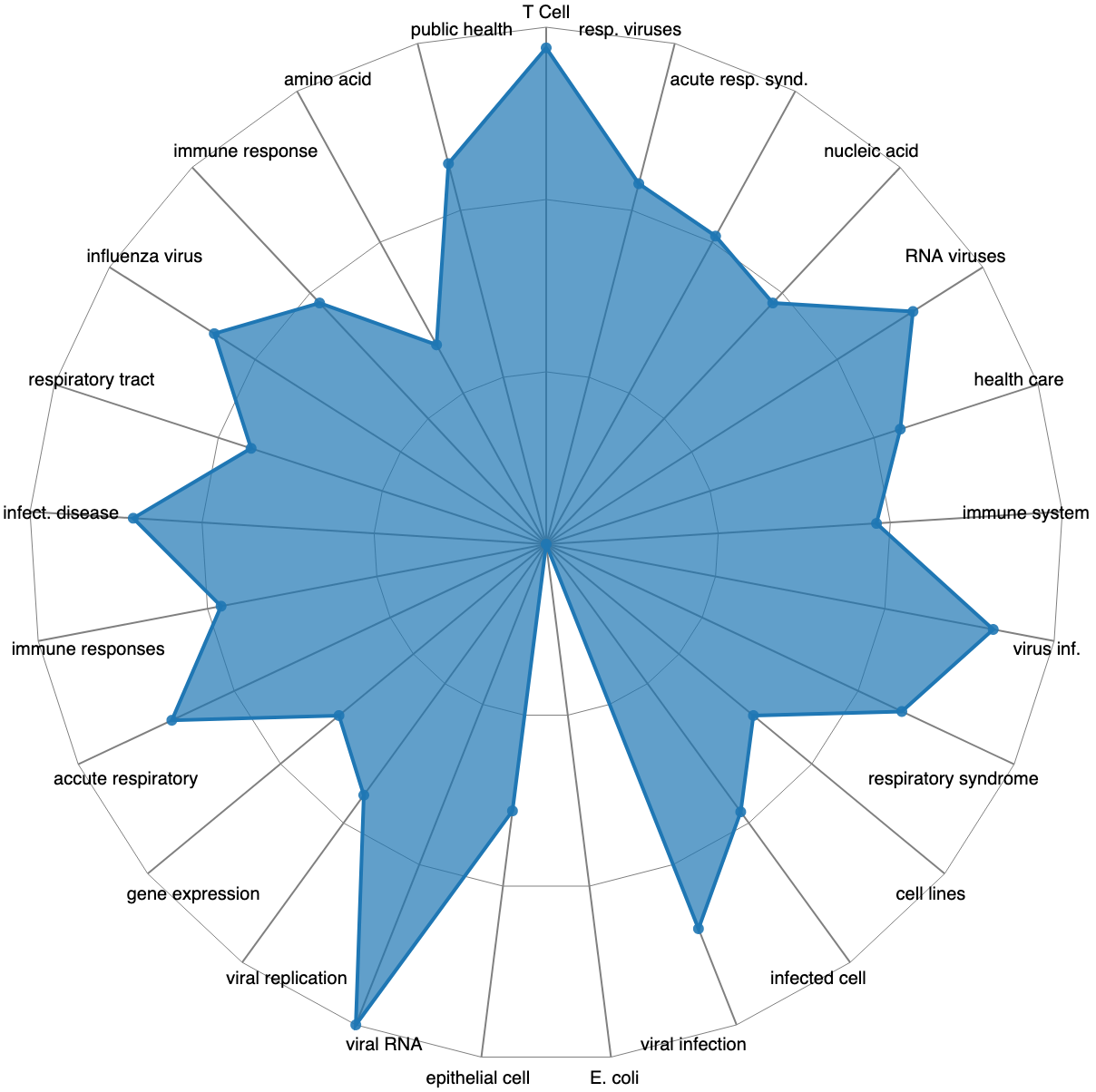}
        \caption{COVID-19 and Malaria}
            \label{fig:comparison_covid_malaria}
    \end{subfigure}
    \begin{subfigure}{.4\linewidth}
        \includegraphics[width=0.9\columnwidth]{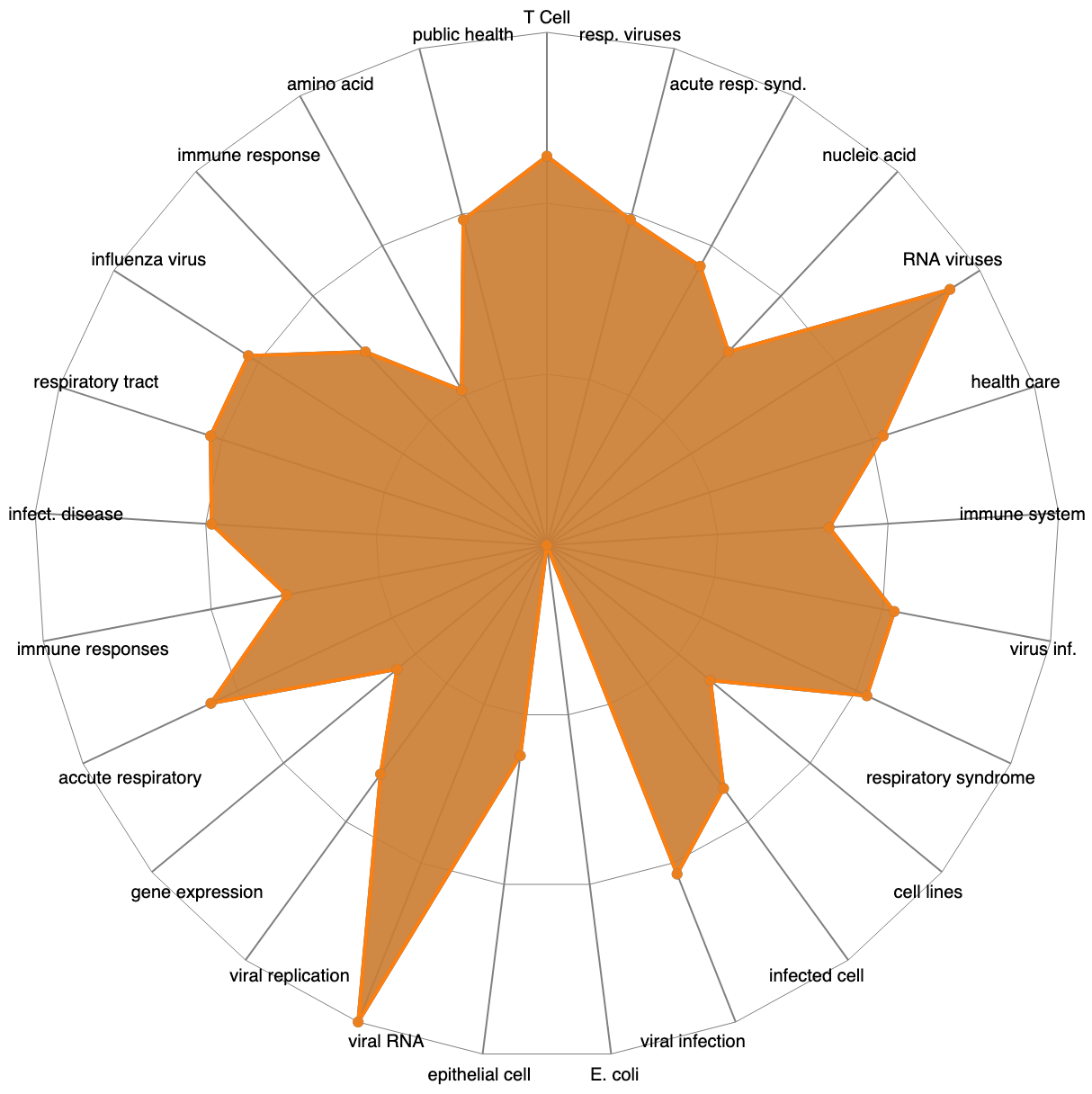}
        \caption{COVID-19 and Conjunctivitis}
            \label{fig:comparison_covid_conjunctivitis}
     \end{subfigure}
\caption{Comparison of diseases based on distances among common terminology described in the embedding spaces created for each of them.}
\label{fig:covid_comparison}
\end{figure}

\begin{itemize}
\item \textbf{Terminology Generation}: The aim of this step was to discover common terminology amongst diseases. We extracted the most frequent terms from a representative sample of 100.000 paragraphs from CORD-19. Given the complex nature of the source texts (full of technical expressions, numbers, figures, symbols)  this task presented a challenge when tested with two terminology extraction methods: Rake \cite{rose2010automatic} and TBXTools\cite{vazquez2018improving}. While the former did not solve many of the above mentioned hindrances and some extracted terms were linguistically incorrect, the latter gave good results that needed only some post-processing work (Table \ref{tab:terms}).
\item \textbf{Representational Model}: Out of the existing state-of-the-art word embedding models, Word2Vec \cite{Mikolov2013} was selected to generate a representational model for each disease. To ensure comparability between models, the same initialization values were used from the pre-trained BioWordVec model \cite{zhang_biowordvec_2019}. Using a proper initialization not only enables direct comparability between disease models, but eases the convergence of the model. 
\item \textbf{Similarity Metric}: Diseases were represented as vectorial models, where each word featured in the training corpus found a corresponding embedding. However, not all words in the resulting vocabulary were equally explanatory of the disease, and thus comparisons between models should not be established using the totality of the embedded words. We retrieved the 25 most frequent terms in CORD-19 with the aim of discovering common terms amongst diseases (Table \ref{tab:terms}). They were used as reference points to compare models that visually represent each disease, considering each term as a node, and the distance between each pair of nodes as edges. Similarity between diseases is obtained by averaging individual distances between the representations of the same term on each disease model. Word Mover's Distance \cite{pmlr-v37-kusnerb15-wmd} was used to measure the distance between terms, as it captures the underlying semantics between the resulting word representations. 
\item \textbf{Space Partitioning}: Disease models were initialized using the same values, meaning that the initial embedding of each term was the same across every disease. During training, these embeddings evolve, or \textit{move} throughout the vectorial space to fit the context constraints from the disease-specific training corpus. Therefore, while the initial representations were equal, the final representations differ, as they also embedded specific information about the disease. However, the resulting representations can still be compared. Pairwise comparison between diseases can be established by measuring the Euclidean distance between the embeddings of same terms on each model. Small distances between term representations mean that they occur in a similar context, thus their final embeddings remain close. Figure \ref{fig:covid_comparison} illustrates pairwise comparison between three diseases. According to the obtained results, COVID-19 is closer to Malaria than to Conjunctivitis, as the cumulative distance of all pairwise term distances is lesser in that case. This functionality was also added to our Bio-API service.
\end{itemize}

%

\section{Drugs4Covid Knowledge Graph}
\label{sec:domainknowledge}

All the work described in previous sections allowed us to create the knowledge to enable the search and navigation over the corpus. To facilitate its (re)use, we created a graph that contains this knowledge. This section describes the development of the vocabulary and details the steps followed to build the Drugs4Covid Knowledge Graph (D4C-KG) by combining the drug and disease annotations inferred from the CORD-19 corpus with external resources.

\subsection{The Drugs4Covid vocabulary}

The vocabulary defined to support the D4C-KG was developed following the LOT methodology \cite{lot2019}, relying heavily in the communication with users (i.e. developers generating annotations and KG developers) and domain experts (i.e. biologists and pharmacists) throughout the ontology development process. The main goals of the vocabulary were to represent and relate some parts of the papers analyzed with biomedical concepts, as obtained with the application of language technologies. 

The process started with a meeting where the potential annotations, based on drugs and diseases,  were analyzed. A first set of requirements was extracted: 1) A paper contains paragraphs and sentences; 2) A paragraph and a sentence can mention chemical substances and disorders; 3) A paper mentions a drug; 4) A drug has active substances, which are a type of chemical substances; 5) A drug treats a disorder; 6) A disease has symptoms. The vocabulary requirements evolved throughout the development of the project.

\begin{figure}[hbt!]
\includegraphics[scale=0.45, ]{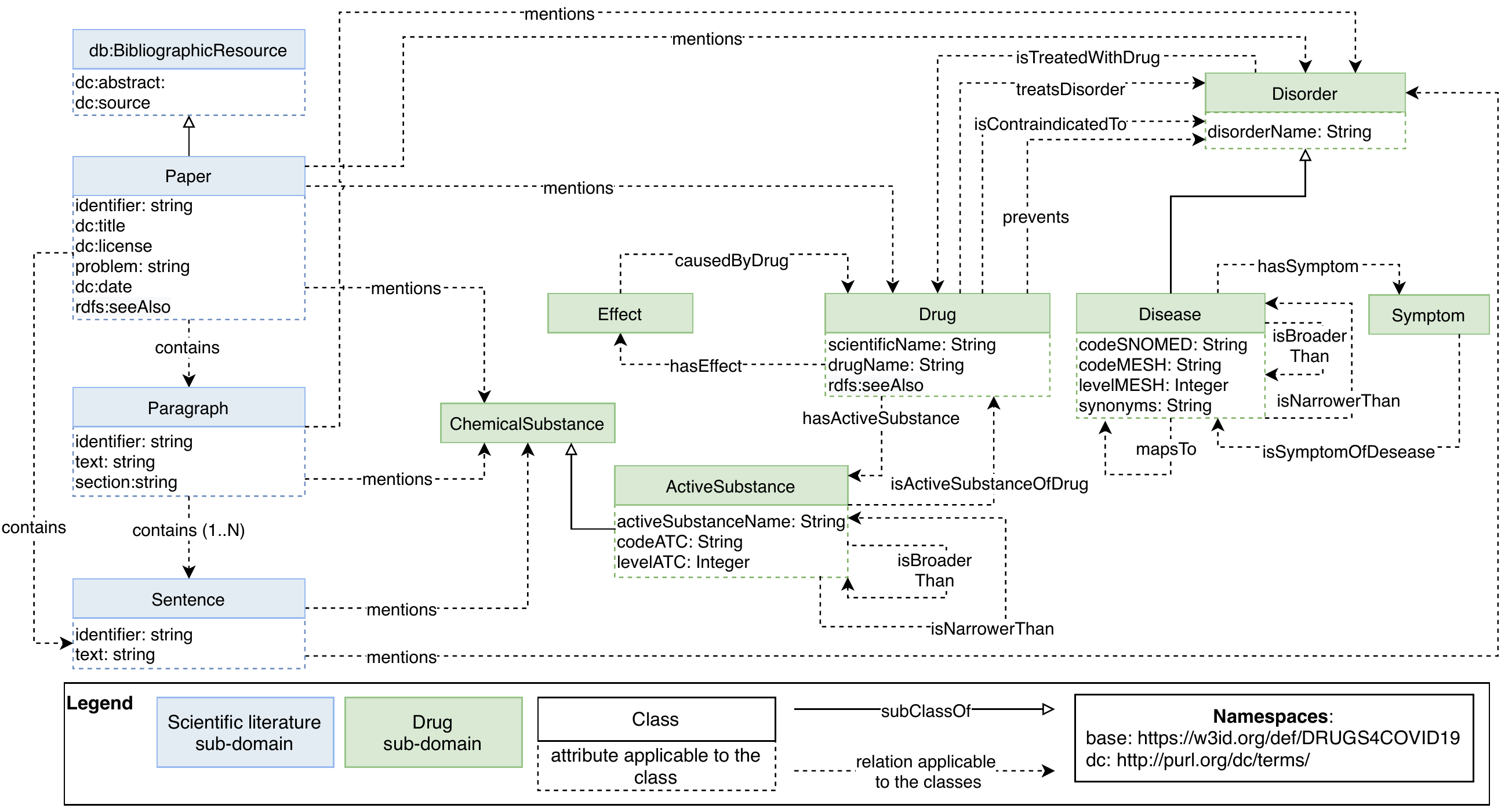}
\centering
\caption{Main classes, properties and attributes of the Drugs4Covid vocabulary. }
\label{fig:vocab}
\end{figure}

The vocabulary conceptualization was carried out following the guidelines for ontology diagrams presented in \cite{garijo2020best} and splitting the team into the two main domains involved in the data, namely ``publications'' and ``biomedical'' domains. Figure \ref{fig:vocab} depicts the vocabulary conceptualization, whose main concepts are: 
\begin{itemize}
    \item \textbf{Paper}: A document of the corpus, subject of annotations. It can contain paragraphs (one of a series of subsections each usually devoted to one idea) and sentences (a sequence of words capable of standing alone to make an assertion, ask a question, or give a command). These concepts can mention disorders and/or chemical substances.
    
    \item \textbf{Disorder}: A disruption to regular bodily structure and function. A disease is often known to be a medical disorder that is associated with specific symptoms, while a symptom refers to a physical or mental feature that is regarded as indicating a condition of disease.
    
    \item \textbf{Chemical substance}: A substance produced by or used in a chemical process. Active substances are chemical substances that are the main ingredients of drugs used to treat or prevent a disorder. Each active substance matches an ATC code as described in the Annotation Section. 
    
\end{itemize}

The vocabulary reuses some Dublin-Core Metadata Terms\footnote{\url{https://dublincore.org/}} for bibliographic resources such as \textit{abstract}, \textit{license}, and \textit{source}. Since we are pursuing a progressively more complex construction of the knowledge graph, we decided to initially create a basic model that allows easy alignment with existing ontologies in the biomedical domain later on. However, some modeling decisions were already taken in this direction, by aligning the active substance class attributes to the ATC annotations from bioontology\footnote{\url{https://bioportal.bioontology.org/ontologies/ATC}}. 

We carried out a twofold evaluation: (1) quality and correctness of the ontology using OOPS!\cite{poveda2014oops}, and (2) a data-directed evaluation where we checked that the data in the indexed corpus can be transformed, annotated and queried with the vocabulary. Finally, the HTML documentation of the vocabulary was generated with Widoco \cite{garijo2017widoco} and published\footnote{\url{https://w3id.org/def/DRUGS4COVID19}} using OnToology \cite{alobaid2018automating}.

\subsection{Knowledge Graph Construction}
The D4C-KG contains RDF annotations extracted from the CORD-19 dataset that are described by the D4C vocabulary. The relations between the source data (annotations) and the developed ontology were declared using the RML mapping language~\cite{DimouSCVMW14}. We used YARRRML\footnote{\url{https://rml.io/yarrrml}}, a human-friendly RML serialization in YAML format, to define 44 subject and 150 predicateObject mappings. The rules created by this technique were easy to define and not time-consuming.

The building process was automated and modular, based on standard W3C technologies in order to ensure the maintainability of the KG construction process. The following steps were performed:
\begin{itemize}
    \item \textbf{Data Preparation}: Annotations from the CORD-19 corpus were exported into CSV files, cleaned and normalized exploiting CSVW annotations.
    \item \textbf{Mapping Definition}: RML mapping rules between the data and the ontology were defined, as well as the rules to link other KGs.
    \item \textbf{RDF Generation}: The cleaned data and mapping rules were processed with SDM-RDFizer~\cite{IJC20} to generate the corresponding RDF dataset.
    \item \textbf{Publication}: The resulting RDF, along with the ATC data available in BioPortal, were published through a Virtuoso SPARQL endpoint\footnote{\url{https://kg.drugs4covid.oeg-upm.net/sparql}}. In addition, federated queries across different datasets can be performed to external knowledge graphs to enhance the completeness of the provided knowledge.
\end{itemize}

\section{Use Cases} \label{sec:usecase}

\begin{figure}[t]
\includegraphics[scale=0.3]{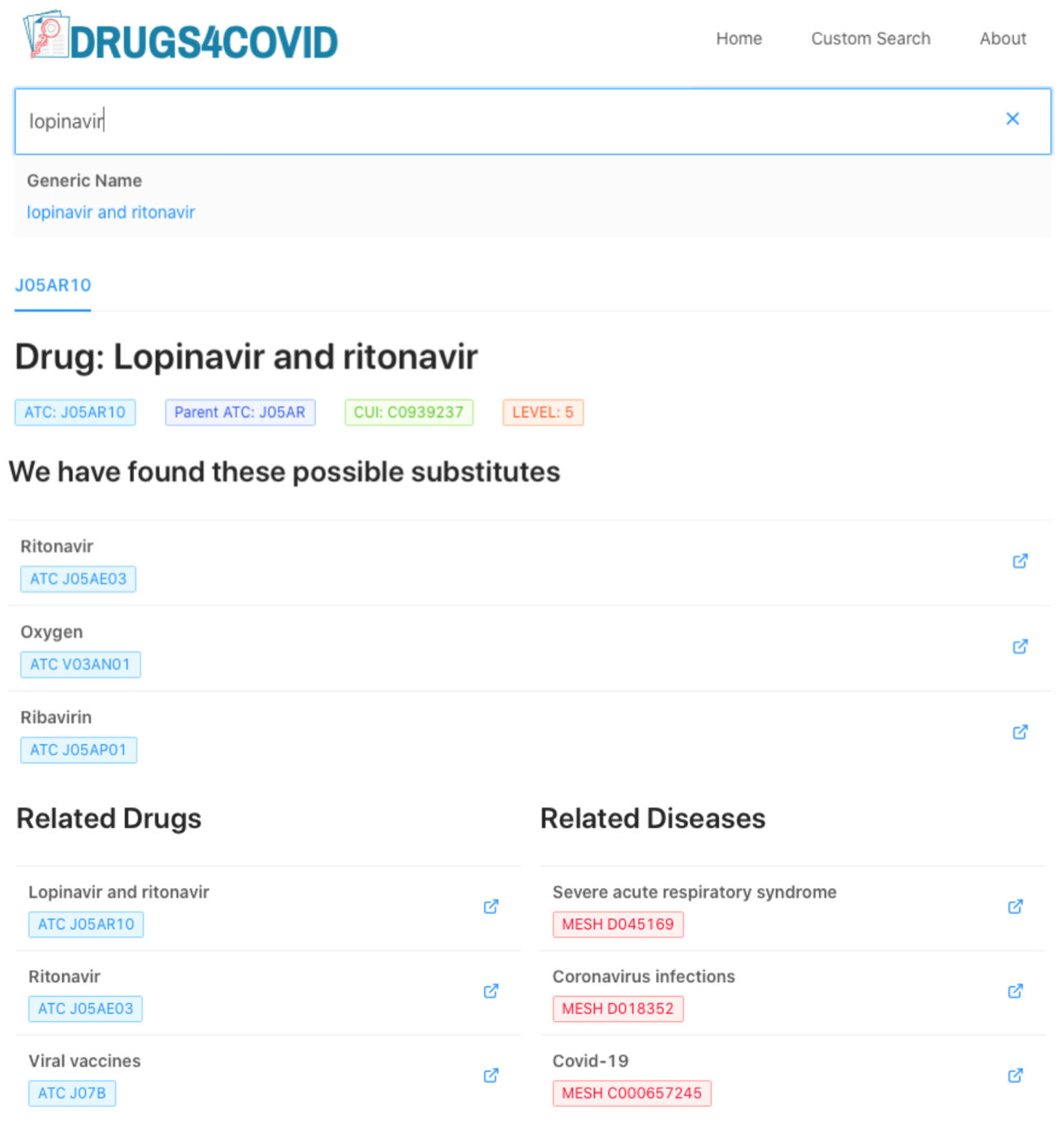}
\centering
\caption{\textbf{Drugs-oriented search interface} showing drugs and diseases related to \textit{lopinavir}.}
\label{fig:search-engine}
\end{figure}

As a result of our work, users can quickly locate the \textit{paragraphs in the articles where immunosuppressive and antimalarial activity with macrolide antibiotics are discussed}; or have a \textit{list of related diseases to mefloquine and azithromycin when used to treat the COVID-19}. 

All this information is available for end users through our search engine, and for anybody who wants to reuse the results through the restful Bio-API and the open knowledge graph described above. The main use scenarios are:

\begin{itemize}
    \item \textbf{Drug-oriented Search}: Our search engine\footnote{\url{https://search.drugs4covid.oeg-upm.net}} provides the paragraphs of the articles where the drug is mentioned, either by its trade name, its generic name or its ATC classification; and a list of related drugs and related diseases inferred from the representational models created for each of them (Fig.\ref{fig:search-engine}).
    \item \textbf{Keyword-oriented Search}: Our RESTful API\footnote{\url{https://search.drugs4covid.oeg-upm.net/customsearch}} browses scientific literature through the relations inferred among drugs\footnote{\url{https://librairy.linkeddata.es/bio-api/drugs}} and diseases\footnote{\url{https://librairy.linkeddata.es/bio-api/diseases}}. The drugs used in conjunction with \textit{lopinavir} (e.g. \url{/bio-api/drugs?keywords=lopinavir}), or the diseases treated with \textit{chloroquine} (e.g. \url{/bio-api/diseases?keywords=chloroquine}) can be easily explored.
    \item \textbf{Pattern-oriented Search}: Our SPARQL endpoint\footnote{\url{https://kg.drugs4covid.oeg-upm.net}} offers a semantic interface for exploration. The most representative/descriptive questions asked by practitioners, pharmacists and general public in natural language were translated into SPARQL.  Listing \ref{lst:sparql} shows one of them. It retrieves the diseases related to drugs that combine immunosuppressant and antimalarial activities with macrolide antibiotics.
\end{itemize}

\begin{lstlisting}[style=SPARQLdeco, 
                  escapechar=|,  
                  caption={List of drugs by a SPARQL query}, 
                  label=lst:sparql]
PREFIX onto: <https://w3id.org/def/DRUGS4COVID19#>

SELECT DISTINCT ?section ?paperTitle ?notation2 ?titleDisease WHERE {
    ?paragraph a onto:Paragraph .
    ?paragraph onto:section ?section .
    ?paper onto:contains ?paragraph .
    ?paper dc:title ?paperTitle .
    ?paragraph onto:mentions ?activeSubstance1 .
    ?paragraph onto:mentions ?activeSubstance2 .
    ?activeSubstance1 skos:notation "P01BA02"^^xsd:string .
    ?activeSubstance2 skos:notation  ?notation2 .
    ?paragraph onto:mentions ?disease .
    ?disease a onto:Disease .
    ?disease onto:MESHCode 'C000657245' .
    ?disease dc:title ?titleDisease .
    FILTER (STRSTARTS(?notation2,"J01FA"))
}
\end{lstlisting}
                                                                


\section{Lessons Learned and Future Work}\label{sec:lessons}

This paper presents our work to create a knowledge graph and a search engine to facilitate the exploration of the CORD-19 corpus through the relations discovered between drugs, diseases and texts. It required an effective collaboration between language technology and semantic web researchers, and with domain experts (doctors and pharmacists) who helped us to resolve doubts and guide functional decisions. The obtained results, software, models and resources are publicly available for anyone to use. 

After the initial release of our services and the creation of a knowledge graph together with its associated vocabulary, we have received numerous requests to continue expanding the types of annotations and the information offered by the knowledge graph. Our results are also being used by third parties, e.g. to obtain "therapeutic targets". The relations sought in this case arise from the proteins described in articles, with the added difficulty that they are usually mentioned by acronyms. Their feedback is promising, and they encourage us to continue incorporating also the management of genes and biological pathways.

This would lead to an increment of the possibilities offered to users when browsing the corpus. Moreover, the exploitation of these annotations in the knowledge graph leads us to increase the reuse of existing biomedical ontologies and terminologies.

We are also working on a Citizen Science initiative with students of Pharmacy and Medicine to validate our annotations and help understand the type of relations that may exist between drugs and diseases. The consensus acquired from the contributions made by these citizen scientists will serve to measure the quality of the results and improve the system as a whole. As in many other Citizen Science projects, a score-based system will be created using parameters such as user participation, quality of contributions (based on gold standards and community contributions), etc. And finally, a helpdesk will allow us receiving  feedback from users to continue improving the results that we provide.

Other more general functionalities that have been already requested by end users are related to usability and to the creation of alerts when a drug, protein or combinations of them appear in the corpus.

\bibliographystyle{splncs04}
\bibliography{biblio}
\end{document}